\documentclass[letterpaper, 10 pt, conference]{ieeeconf}  

\IEEEoverridecommandlockouts                              

\usepackage{amsmath} 
\usepackage{amssymb}  
\usepackage{graphicx}
\usepackage{soul}
\usepackage{xcolor}
\usepackage[acronym]{glossaries}
\usepackage[ruled,vlined,noend]{algorithm2e}
\usepackage{algpseudocode}%
\usepackage{multicol}
\usepackage{cite}
\usepackage{bm}
\usepackage{tabularray}

\title{\LARGE \bf
A Surface Adaptive First-Look Inspection Planner for Autonomous Remote Sensing of Open-Pit Mines
}

\author{Vignesh Kottayam Viswanathan, Vidya Sumathy, Christoforos Kanellakis and George Nikolakopoulos
\thanks{The authors are with Robotics and AI, Luleå University of Technology 97187, Luleå,Sweden
        {\tt\small \{vigkot, visdum, chrkan and geonik\}@ltu.se}}%
\thanks{
The project is funded by the European Union's Horizon Europe Research and Innovation Programme, under the Grant Agreement No. 101091462 m4mining. The authors would like to thank the Feiring AS company for providing access to their site and the m4mining partners for the data collection missions.}
\thanks{The authors would like to thank Ilias Tevetzidis in contributing as a safety-pilot for the experiment trials.}
}

\begin{document}

\maketitle
\thispagestyle{empty}
\pagestyle{empty}

\begin{abstract}

In this work, we present an autonomous inspection framework for remote sensing tasks in active open-pit mines. Specifically, the contributions are focused towards developing a methodology where an initial approximate operator-defined inspection plan is exploited by an online view-planner to predict an inspection path that can adapt to changes in the current mine-face morphology caused by route mining activities. The proposed inspection framework leverages instantaneous 3D LiDAR and localization measurements coupled with modelled sensor footprint for view-planning satisfying desired viewing and photogrammetric conditions. The efficacy of the proposed framework has been demonstrated through simulation in Feiring-Bruk open-pit mine environment and hardware-based outdoor experimental trials. The video showcasing the performance of the proposed work can be found here: https://youtu.be/uWWbDfoBvFc
\end{abstract}

\section{Introduction}

In recent years, there has been a significant increase in the deployment of autonomous robots for large-scale environment surveying and mapping~\cite{kanellakis2018towards,jung2019toward,selin2019efficient}. Specifically, the robotization of remote monitoring operations in the mining industry is currently pursued with significant interest~\cite{shahbazi2015development,tong2015integration,battulwar2020practical,wajs2021modern}. Micro Aerial Vehicles (MAVs) equipped with specialized sensors (e.g. hyper-spectral imaging, gas sensing, etc) can access remote or inaccessible areas of mining sites that may be challenging or dangerous for humans to reach (e.g. during the blasting cycle)~\cite{kersnovski2017uav,bamford2020continuous}. This capability enables comprehensive monitoring and rapid surveying vast areas of mine site and can provide real-time data. The autonomous deployment of MAVs in open-pit mines has limitation including the mine face evolution, predefined waypoint selection and the lack of mission design on data quality measures.Among the fundamental capabilities that are required to enable the autonomous deployment of aerial robotic systems in such environments is to detect the surface of interest that should be sensor scanned, generate reactively collision-free inspection way-points along the mine face, while also considering the sensor footprint to establish the required data quality (e.g. photogrammetric footprint for visual 2-dimensional cameras).

To address this challenge, the contributions of this article focus on the development of a surface adaptive path planning framework, based on the First-Look inspection planning method~\cite{viswanathan2022first}. More specifically, the contribution stems from the novel mission design applicable for varying surface morphology during inspection runs. In the proposed work, we present an online inspection planner capable of leveraging operator-defined initial inspection route. While the operator designs the flight path according to previously collected information, the proposed framework expands the First-Look scheme using instantaneous 3D pointcloud measurements from the onboard LiDAR sensor to actively track the closest area through a KDTree search to adapt the operator's path to the current surface morphology. To attain the desired quality of imaging during inspection, the proposed planner considers modelled sensor footprint and required viewing distance, which based on the selected planning horizon, dynamically and recursively estimates the future sequence of view-poses with respect to the operational parameters.

In view of related works for model-based inspection planning direction,~\cite{mansouri2018cooperative} present an optimal coverage path planner given an apriori known infrastructure model. In this approach, the 3D model is processed to generate candidate view-points in an offline manner. The inspection plan is then tracked as-is by the aerial platform. In~\cite{jing2019coverage}, the authors present a model-based offline view-planner for inspecting 3D infrastructures. The discussed approach introduces a sampling-based planning problem which refines over an initially generated waypoints based on visibility, travel cost and viewing requirements given the topological information.~\cite{biundini2021framework} addresses model-based inspection planning with a two-step process. In this work, the overall inspection route is designed offline subject to mission time, viewing criteria and the processed 3D model. Subsequently, the aerial platform is tasked to track with an objective to recognize interesting regions and execute a closer pre-defined inspection route for collecting detailed information of the structure. In~\cite{viswanathan2022experimental}, the authors present an offline view-planning policy for inspection in constrained environments. In this approach, the inspection route is defined by processing the 3D model, environment knowledge and required viewing conditions.

Considering online view-planning for unknown environments,~\cite{song2017online} presents the task of modelling unknown environments through next best view (NBV) planning approach. The authors demonstrate a optimal view-planning policy based on imposed sensing requirements to generate 3D models of apriori unknown environments. In~\cite{hepp2018plan3d}, the authors present an explore-then-exploit strategy which functions to generate a coarse 3D map of the environment from an initial flight. This is the exploited to generate view-poses to collect detailed and dense information of the environment. In~\cite{song2018surface}, the authors propose a method that balances exploration efficiency from a volumetric perspective with considerations for the quality of the observed surface.~\cite{schmid2020efficient} introduces an online path planning algorithm that utilizes sampling to efficiently explore a 3D environment. It leverages a single Rapidly-exploring Random Tree (RRT) to achieve complete coverage of the environment. 

Compared to the current-state-of-art, the proposed framework utilizes the high-level operator-defined inspection route to focus the effort of the online view-planner. Unlike model-based methods that assume static environments where deviations in the 3D model at hand and in-situ are assumed to be negligible, the situation is different for an active mine site. In such cases, the mine-face recedes due to regular operations and thus offline planning may not entirely reflect real conditions. The proposed framework dynamically adapts to changed environment conditions by replanning view references based on real-time observations. Additionally, it avoids the exhaustive coverage employed by non-model-based methods by focusing inspection efforts on the relevant region as designated by the operator.

Thus, the technical aspects of the contributions can be summarized as follows:

\begin{enumerate}
    \item  We present a novel point-and-click inspection autonomy for autonomous remote sensing of open-pit mine environments. The proposed online inspection planner dynamically adapts the required view-poses based on instantaneous 3D LiDAR measurements and modelled sensor footprints, while maintaining awareness of operator-defined high-level inspection route.

    \item We demonstrate the efficay of the proposed framework through large-scale simulated point and click mission in Feiring-Bruk open-pit mine environment. Additionally, we present experimental evaluations of the inspection framework in an outdoor scenario on an aerial platform.
\end{enumerate}

The rest of the article is structured as it follows. In Section~\ref{sec:prop_meth} the proposed online inspection planner is being presented, while in Section~\ref{sec:setup} details on the established setup of the simulation environment and the hardware for experimental evaluations are explained. Section~\ref{sec:results} presents the results and corresponding discussions of the performance of the proposed framework while Section~\ref{sec:conclusions} outlines the drawn conclusions as well as future works.



\section{Proposed Methodology}\label{sec:prop_meth}

\begin{figure}[htpb]
    \centering
    \includegraphics[width = \linewidth]{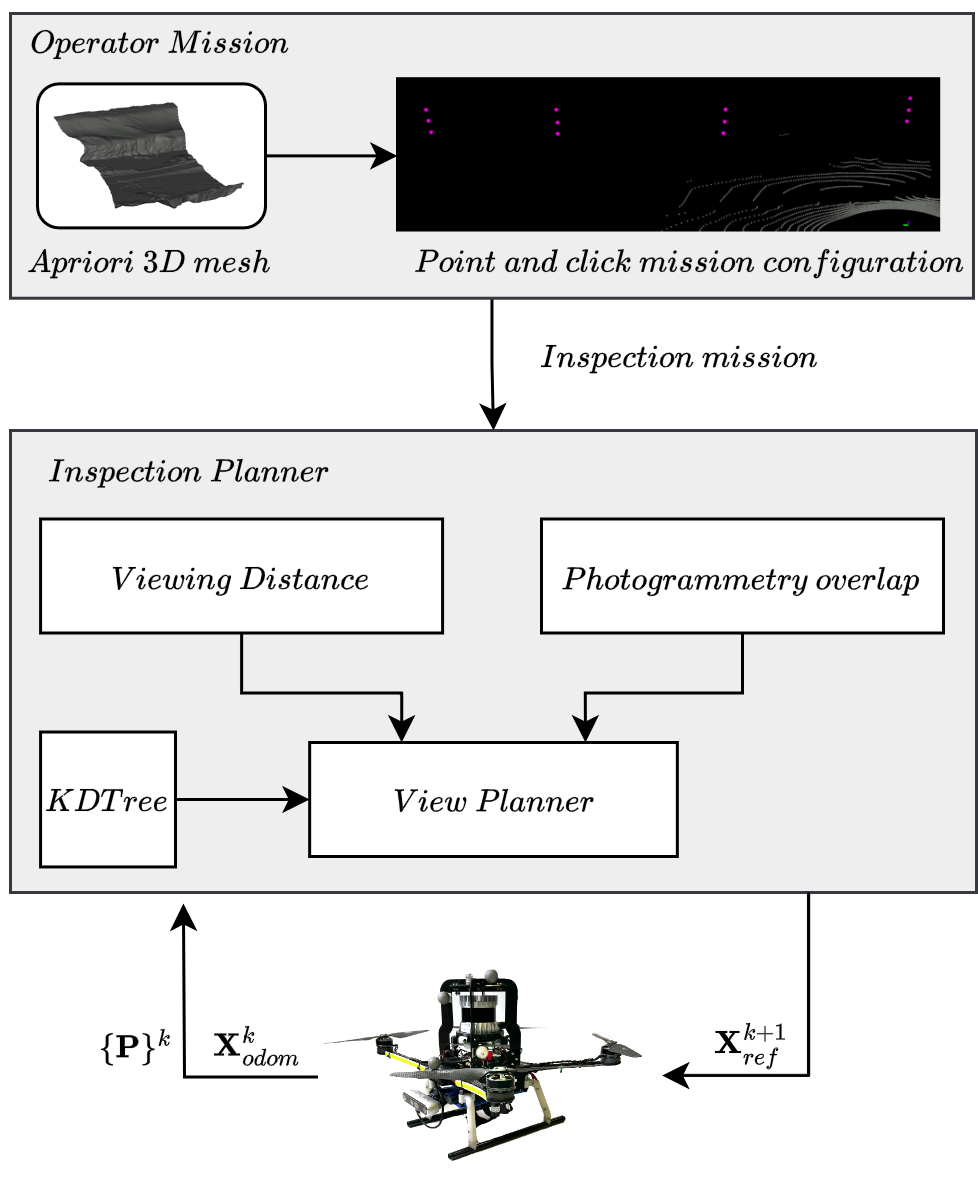}
    \caption{A functional representation of the evaluated mission architecture for open-pit mine face inspection.}
    \label{fig:framework}
\end{figure}

Figure.~\ref{fig:framework} presents an overview of the proposed framework for open-pit mine face inspection. Our work proposes a two-layered framework for an active open-pit mine-face inspection. The first layer utilizes prior information shared as an operator-defined initial inspection route based on available 3D model of the environment. The second layer consists of online view-planner which dynamically refines and replans the required viewing references based on real-time sensor measurements, ensuring alignment with the current environment conditions. Primarily, the framework is built on top of the elements initially introduced in the First-Look Inspection framework~\cite{viswanathan2022first}. Compared to~\cite{viswanathan2022first}, the proposed planner exploits instantaneous 3D LiDAR point-cloud ($\{\textbf{P}\}^k \in \mathbb{R}^3$) and localization measurements ($\textbf{X}_{odom}^k = [\textbf{X}_{odom[x,y,z]}^k,\textbf{X}_{odom[\phi,\theta,\psi]}^k] \in \mathbb{R}^3\times \mathbb{SO}(3)$) to generate reference view-pose ($\textbf{X}_{ref}^{k+1} = [\textbf{X}_{ref[x,y,z]}^{k+1},\textbf{X}_{ref[\phi,\theta,\psi]}^{k+1}] \in \mathbb{R}^3\times \mathbb{SO}(3)$) to be maintained.

\subsection{First-Look Inspection Planner}

The First-look inspection planner operates on instantaneous 3D LiDAR measurements to generate the reference view-pose to be maintained by the aerial platform during inspection. The view-pose generation is subject to the definition of two main operational parameters, \textit{viewing distance} ($d_{view} \in \mathbb{R}^+$) and desired \textit{photogrammetric} overlap. We denote $\Vec{\bm{\nu}}^k_x,\Vec{\bm{\nu}}^k_y,\Vec{\bm{\nu}}^k_z \in \mathbb{R}^3$ as unit vectors capturing the ego orientation of the aerial platform at each instant along $X,Y,Z$ axes respectively (refer Fig.~\ref{eqn:overlap}). 

The advantage of deploying the First-look planner is in its capability to generate profile-adaptive view-pose which allows the aerial platform to follow the contour of the locally observed surface during inspection. This is achieved through the continuous evaluation of the desired ego orientation of the platform for projecting the next view-pose. We denote $\textbf{P}_{nn}^k \in \mathbb{R}^3 \subseteq \{\textbf{P}\}^k$ as the nearest neighbour 3D point of the locally observed mine-face extracted through a KDTree search on the perceived point-cloud measurement from $\textbf{X}_{odom[x,y,z]}^k$. Given $\textbf{P}_{nn}^k$ and $\textbf{X}_{odom[x,y,z]}^k$, the unit-vectors are formulated as in~\eqref{eqn:unit_vec},

\begin{equation}\label{eqn:unit_vec}
\begin{aligned}
    \Vec{\bm{\nu}}_x^k &= \frac{\textbf{P}_{nn}^k - \textbf{X}_{odom[x,y,z]}^k}{||\textbf{P}_{nn}^k - \textbf{X}_{odom[x,y,z]}^k||} \\
    \Vec{\bm{\nu}}_y^k &= \Vec{\bm{\nu}}_{up}\times  \Vec{\bm{\nu}}_x^k  \\
    \Vec{\bm{\nu}}_z^k &= \Vec{\bm{\nu}}_x^k \times \Vec{\bm{\nu}}_y^k 
\end{aligned}
\end{equation}

with $\Vec{\bm{\nu}}_{up} = [0,0,1] $ defined as the unit pointing vector along the Z-axis of the aerial platform.
\begin{figure}[htpb]
    \centering
    \includegraphics[width = 0.9\linewidth]{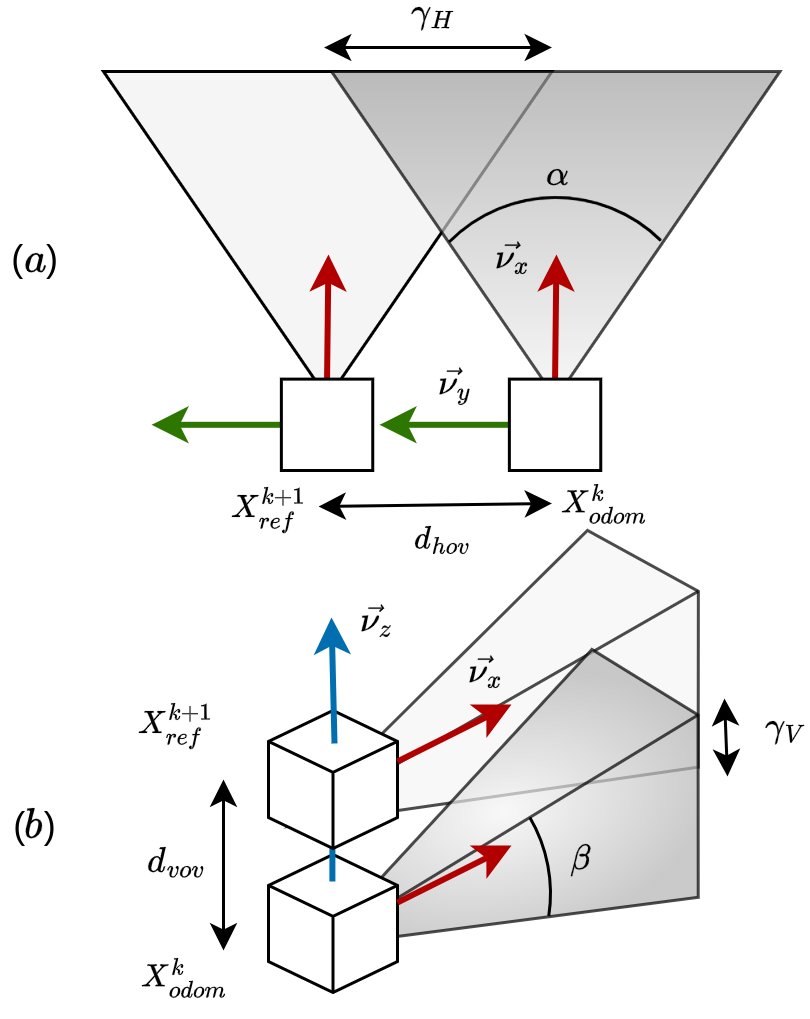}
    \caption{A graphical representation of the modelled photogrammetric characteristics utilized for horizontal (Fig.~\ref{fig:cam_charac}(a)) and vertical overlap (Fig.~\ref{fig:cam_charac}(b)).}
    \label{fig:cam_charac}
\end{figure}

Equation~\eqref{eqn:overlap} presents the mathematical formulation of extracting the required absolute distance to satisfy the desired horizontal and vertical overlap conditions based on the respective field-of-view of the onboard camera. Let $d_{hov},d_{vov} \in \mathbb{R}^+$ denote the distance of overlap required by the planner in order to maintain photogrammetric conditions between two consecutive view-pose. Figure.~\ref{fig:cam_charac} provides a visual description of the modelled photogrammetric constraints  in~\eqref{eqn:overlap}.

\begin{equation}\label{eqn:overlap}
\begin{aligned}
    d_{hov} &= 2\tan(\frac{\alpha}{2})|\textbf{P}_{nn}^k - \textbf{X}_{odom[x,y,z]}^k|| \\& - 2\tan(\frac{\alpha}{2})||\textbf{P}_{nn}^k - \textbf{X}_{odom[x,y,z]}^k||\gamma_H \\
    d_{vov} &= 2\tan(\frac{\beta}{2})||\textbf{P}_{nn}^k - \textbf{X}_{odom[x,y,z]}^k|| \\& - 2\tan(\frac{\beta}{2})||\textbf{P}_{nn}^k - \textbf{X}_{odom[x,y,z]}^k||\gamma_V
\end{aligned}
\end{equation}

where $\alpha,\beta \in \mathbb{R}^+$ are the physical characteristics of the onboard camera for horizontal and vertical field of views and $\gamma_{H},\gamma_{V}\in \mathbb{R}^+$ denote the desired horizontal and vertical overlap respectively.

Finally, to maintain the viewing distance during inspection we find the signed difference between the desired ($d_{view}$) and the current viewing distance ($||\textbf{P}_{nn}^k - \textbf{X}_{odom[x,y,z]}^k||$), where $||\cdot||$ denote taking the norm. The required deviation to be maintained along the viewing direction to satisfy inspection constraints is modelled as in~\eqref{eqn:view_dist},

\begin{equation}\label{eqn:view_dist}
    d_{insp} = ||\textbf{P}_{nn}^k - \textbf{X}_{odom[x,y,z]}^k|| - d_{view}
\end{equation}

Thus, the mathematical backbone of the \textbf{View-Planner} module (refer~Fig.\ref{fig:framework}) to generate reference inspection view-pose is then defined as in~\eqref{eqn:main_flip}.

\begin{subequations}\label{eqn:main_flip}
\begin{equation}
     \textbf{X}_{ref[x,y,z]}^{k+1} =  \textbf{X}_{odom[x,y,z]}^k + \Vec{\bm{\nu}}_x^k d_{insp} + \Vec{\bm{\nu}}_y^k d_{hov} + \Vec{\bm{\nu}}_z^k d_{vov} 
\end{equation}
\begin{equation}
     \textbf{X}_{ref[\psi]}^{k+1} =   \arctan(\Vec{\bm{\nu}}_x^{k+1}(1),\Vec{\bm{\nu}}_x^{k+1}(0))
\end{equation}
\end{subequations}

where, $\Vec{\bm{\nu}}_x^{k+1}$ is formulated as follows with $\textbf{P}_{nn}^{k+1} \subseteq \{\textbf{P}\}^k$ being reevaluated for $\textbf{X}_{ref[x,y,z]}^{k+1}$,

\begin{subequations}
\begin{equation}
    \Vec{\bm{\nu}}_x^{k+1} = \frac{\textbf{P}_{nn}^{k+1} - \textbf{X}_{ref[x,y,z]}^{k+1}}{||\textbf{P}_{nn}^{k+1} - \textbf{X}_{ref[x,y,z]}^{k+1}||} \nonumber
\end{equation}
\end{subequations}

Since the onboard camera model is forward-facing (refer Fig.~\ref{fig:cam_charac}) and fixed on the aerial platform, we only generate yaw attitude reference ($\textbf{X}_{ref[\psi]}^{k+1}$) and consider the roll and pitch references ($\textbf{X}_{ref[\phi,\theta]}^{k+1}$) to be null.

\begin{figure}[htpb]
    \centering
    \includegraphics[width = \linewidth]{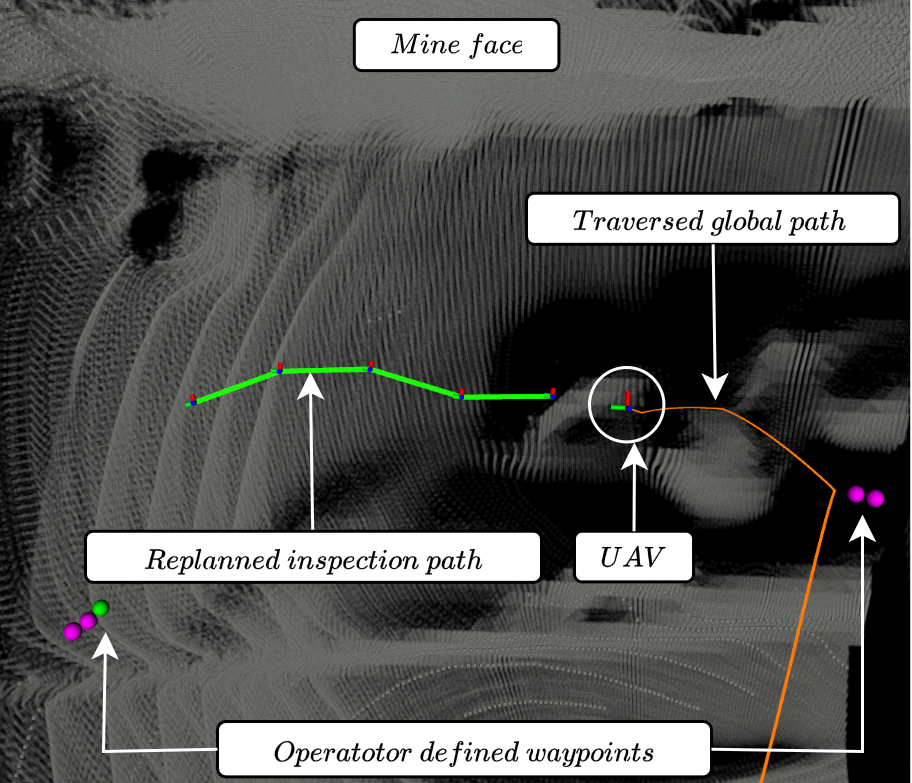}
    \caption{A run-time snapshot of the predicted inspection path (in \textit{green}) being generated during the simulation of mine-face inspection of the Feiring-Bruk open-pit mine.}
    \label{fig:flip_pred}
\end{figure}

This works extends the capability of the \textbf{View-Planner} module by enabling the autonomy to predict the evolution of the inspection path $\Pi_{ref} = \{\textbf{X}_{ref}^{k+1},\textbf{X}_{ref}^{k+2},...,\textbf{X}_{ref}^{k+N}\}$ over a desired horizon $N \in \mathbb{Z}^+$ based on $k^{th}$ instant observations. The predictive component within the proposed framework leverages~\eqref{eqn:main_flip} and $\{\textbf{P}\}^k$ to output a potential evolution of the inspection path, based on the current point-cloud observation. Figure.~\ref{fig:flip_pred} visualizes the functionality of the integrated predictive component. The predicted path is shown as a bold \textit{green} line. At step $k+1$, the autonomy presumes the generated reference $\textbf{X}_{ref}^{k+1}$ at step $k$ as it's localization measurement and replans using~\eqref{eqn:unit_vec}-\eqref{eqn:main_flip} and the corresponding $\textbf{P}_{nn}^{k+2}$ measurement for $\textbf{X}_{ref}^{k+2}$. This process is recursively calculated over the defined planning horizon $N$ to generate the predicted inspection path and is continuously reevaluated as the aerial platform tracks the required references over mission duration.

\section{Setup and Evaluation}\label{sec:setup}

We evaluate the proposed framework onboard an Unmanned Aerial Vehicle (UAV) equipped with Realsense D435 stereo-camera, Inertial Measurement Unit (IMU) and 3D LiDAR in both simulation and hardware evaluations. We use the RotorS~\cite{Furrer2016} simulator along with Gazebo and ROS Noetic on Ubuntu 20.04 LTS operating system with i9-13900K CPU and 128 GB RAM. For experimental evaluations, we use an Intel NUC i5 8365U computational board running ROS Noetic and Ubuntu 20.04 operating system. Voxblox~\cite{oleynikova2017voxblox} package is used in this work for 3D reconstruction of the mine-face.

\begin{figure}[htpb]
    \centering
    \includegraphics[width = \linewidth]{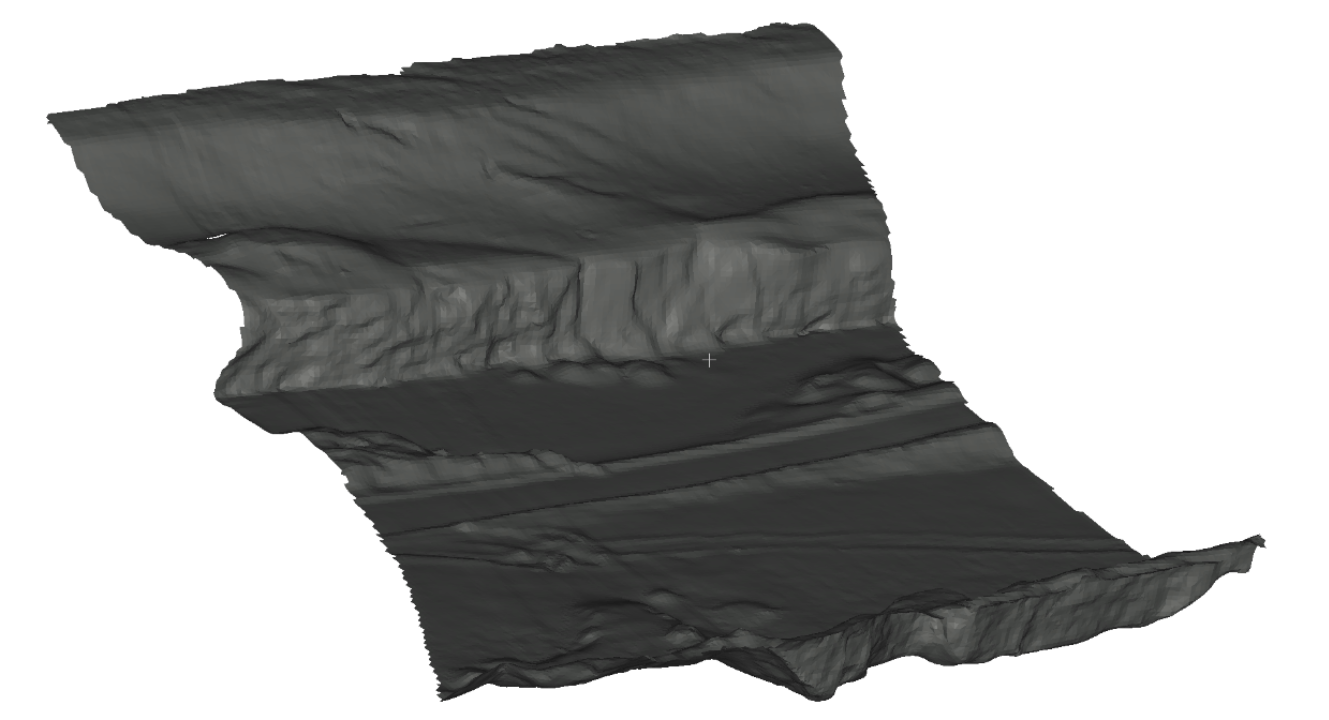}
    \caption{A visual presentation of the 3D simulation environment based on the open-pit Feiring-Bruk mine for evaluation of the proposed inspection framework.}
    \label{fig:mine_mesh}
\end{figure}

We utilize the Feiring-Bruk open-pit mine environment shown in Fig.~\ref{fig:mine_mesh} to run the simulation scenario in GAZEBO. The implemented scenario simulates an active mine-face inspection. The mission starts with an operator defining high-level way-points along the mine-face based on historical 3D mesh collected in previous run. Once initialized, the inspection planner replans the commanded the inspection route based on current environment observation and  viewing requirements. The planner uses the operator-defined waypoints as landmarks and updates them as it passes through the locality of each during the mission. Once the planner has reached the locality of the last defined landmark, inspection is terminated and the UAV is commanded back to the starting position.

Table~\ref{tab:sim_params} presents the simulation parameters taken into consideration for the simulated inspection scenario of the Feiring-Bruk mine. In order to ensure tracking of reference pose required by the planner, we implement Nonlinear Model Predictive Control (NMPC) policy discussed in~\cite{lindqvist2020nonlinear}. The authors kindly direct the readers to the reference for more detailed information. Once $\Pi_{ref}$ is obtained at the end of first iteration of planning, the reference for the next immediate instant $\textbf{X}_{ref}^{k+1} \subseteq \Pi_{ref}$ is fed to the onboard controller for tracking.

\begin{table}
\centering
\caption{A tabular representation of the parameters used in the simulation evaluation of open-pit mine face inspection.}
\label{tab:sim_params}
\begin{tblr}{
  hline{1,8} = {-}{0.08em},
  hline{2} = {-}{0.05em},
}
\textit{Parameters} & \textit{Value} \\
d$_{view}$         & 20 m           \\
$\alpha$              & $69.4^\circ$           \\
$\beta$               & $45^\circ$             \\
$\gamma_{H}$       & 0.8            \\
$\gamma_{V}$       & 0.8            \\
$N$                & 5
\end{tblr}
\end{table}

During experimental evaluations, we additionally supplement the low-level autonomy with Artificial Potential Field (APF) guided collision avoidance scheme based on instantaneous 3D LiDAR measurements. A comprehensive discussion on the utilized control methodology can be found in~\cite{lindqvist2022adaptive}. 

\section{Results and Discussion}\label{sec:results}

\begin{figure*}
    \centering
    \includegraphics[width = \linewidth]{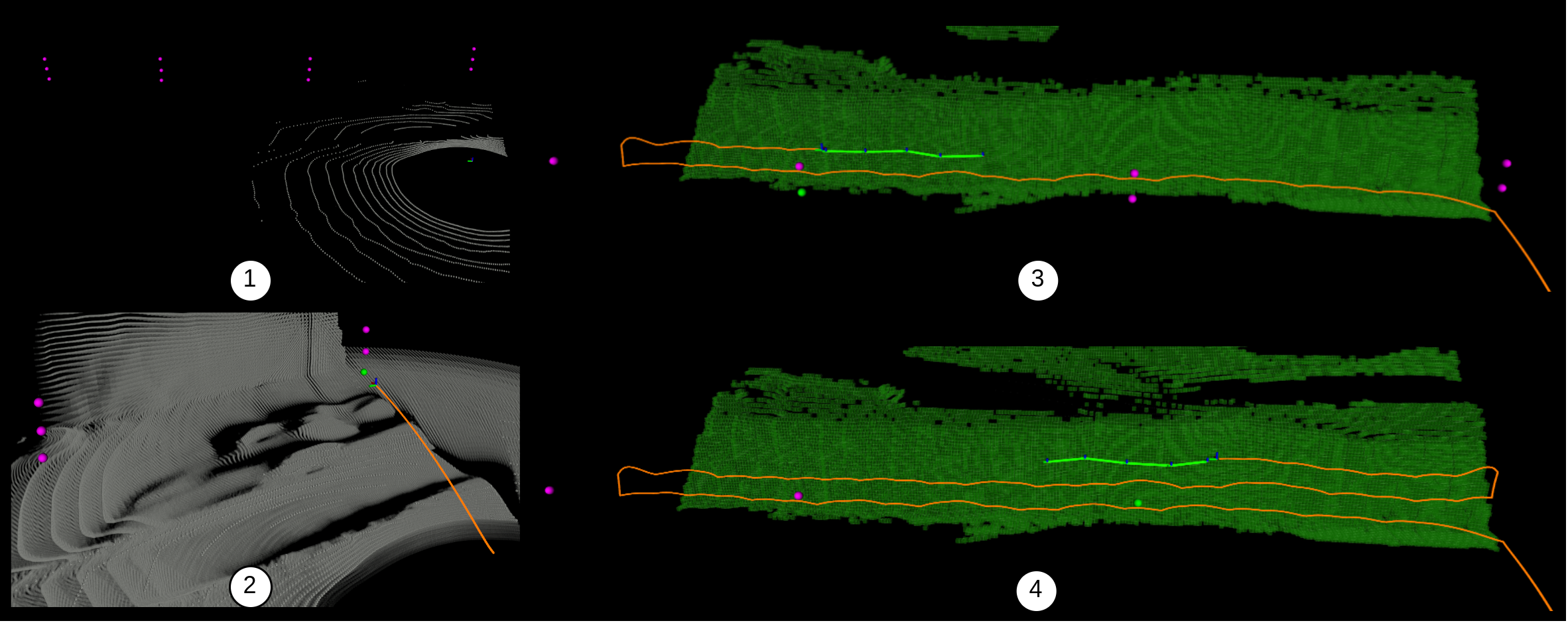}
    \caption{A collage composed of run-time snapshots taken during the simulated run at the Feiring-Bruk mine. Fig.~\ref{fig:sim_collage}(1) presents the initial operator-defined high-level reference waypoints indicated as \textit{maroon} spherical markers. Fig.~\ref{fig:sim_collage}(2) captures the aerial platform en-route to the first waypoint marker upon mission initialization. The traversed route of the UAV is shown via bold \textit{orange} line. Figures.~\ref{fig:sim_collage}(3)-(4) capture the behaviour of the aerial platform during inspection of the open-pit mine face. The observed regions of the mine-face during inspection are shown as \textit{green} markers.}
    \label{fig:sim_collage}
\end{figure*}

\begin{figure*}
    \centering
    \includegraphics[width = \linewidth]{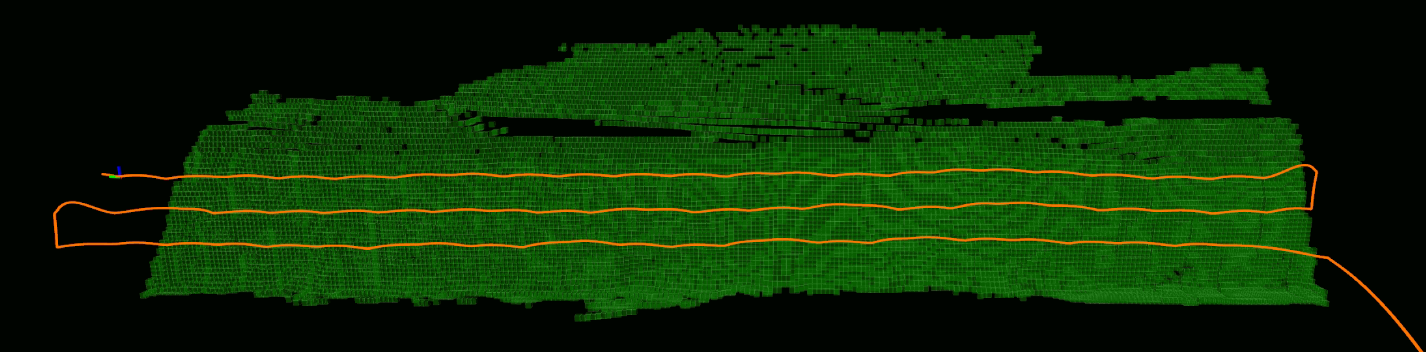}
    \caption{An overview of the traversed path at the end of mine-face inspection mission. The route (in \textit{orange}) is overlayed against the observed regions of the mine-face (in \textit{green}) reconstructed using Voxblox~\cite{oleynikova2017voxblox}.}
    \label{fig:sim_final_route}
\end{figure*}

Figure.~\ref{fig:sim_collage} presents a composite representation of the demonstrated open-pit mine face inspection with instances captured at several stages during the mission. Figure.~\ref{fig:sim_collage}(1) portrays the high-level advisory route fed to the planner indicated as \textit{pink} spherical markers. The 3D way-points defining the route are established through a Point and Click interaction via Rviz Graphical User Interface (GUI). Once the initial inspection route is defined, the mission is triggered. Figure.~\ref{fig:sim_collage}(2) captures the UAV travelling towards the first way-point defined to start inspection of mine-face upon triggering the mission. In the figure, the path taken by the UAV is depicted as an \textit{orange} line. Subsequent to arriving at the initial way-point, the autonomy adaptively plans to position the UAV satisfying the imposed inspection characteristics. Figures.~\ref{fig:sim_collage}(3)-(4) capture the traversed route during inspection of the mine-face. The reconstructed mine-face from 3D point-clouds during inspection is represented as \textit{green} voxels.

Figure.~\ref{fig:sim_final_route} presents the traversed route during inspection of the mine-face overlaid against the observed mine-face regions. As can be inferred, the planner generated route outlines the one defined initially by the operator but also exhibits reactive local planning based instantaneous observations to maintain the desired viewing distance and photogrammetric overlap. Thus, for situations where the active mine-face for inspection can recede farther back due to day-to-day operations, the proposed framework is capable to adapt to environmental condition at hand.

\begin{figure}[htpb]
    \centering
    \includegraphics[width = \linewidth]{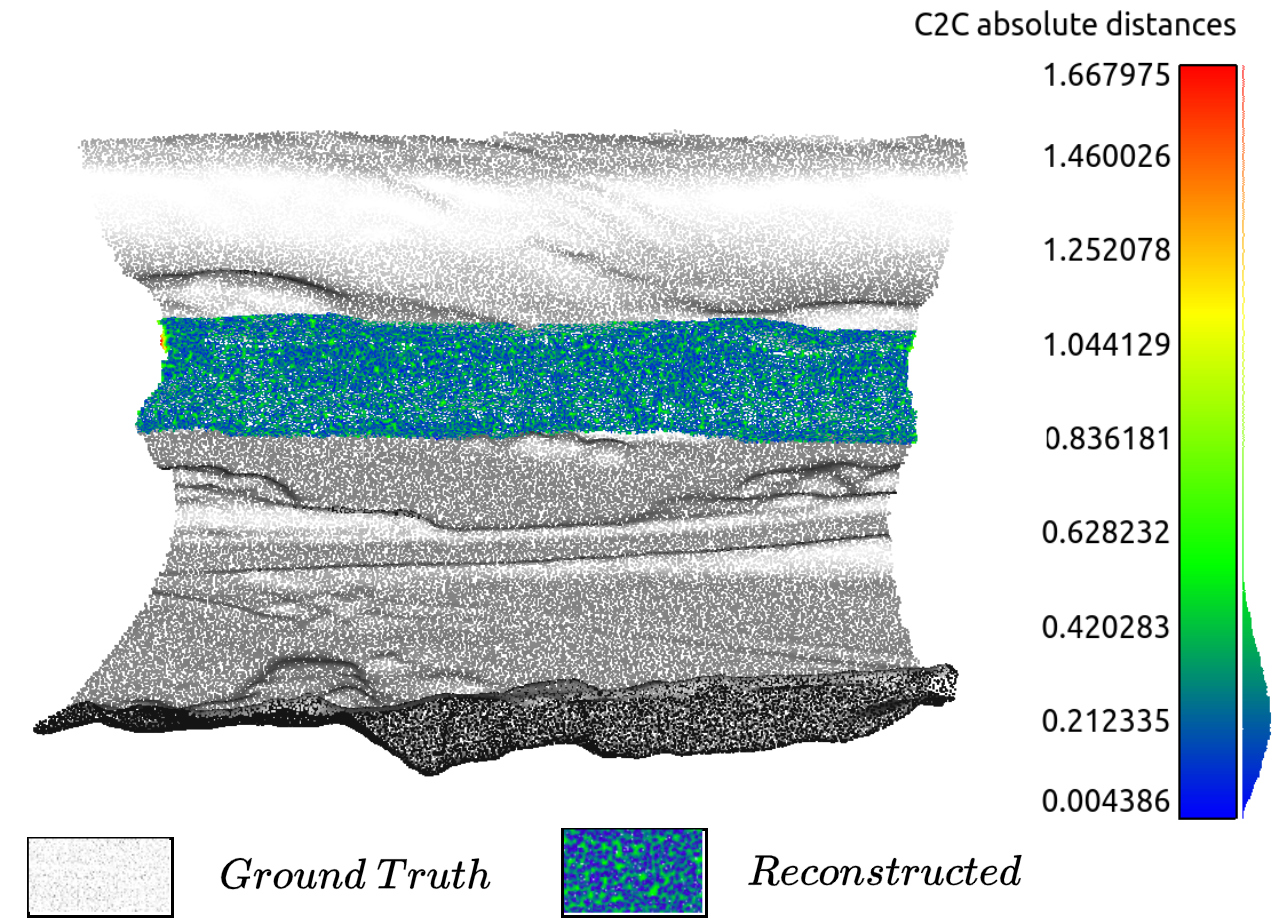}
    \caption{A qualitative comparison between the reference ground-truth point-cloud representation (in \textit{grey}) of the mine-face and the one obtained after inspection (in \textit{blue-green}). The color scale provided in the right provides the absolute distance error of the registered point-cloud against the reference.}
    \label{fig:c2c_quali}
\end{figure}

Figure.~\ref{fig:c2c_quali} presents a qualitative analysis of the reconstructed mine-face obtained from inspection against the ground-truth mesh information. The point-cloud representations were down-sampled to 1000000 points. Employing a cloud-to-cloud comparison in CloudCompare, the reconstructed point-cloud has a mean distance error of 0.256 m when compared with the reference with a maximum outlier error of 1.667 m registered in the mesh fringes.

\begin{figure}[htpb]
    \centering
    \includegraphics[width = \linewidth]{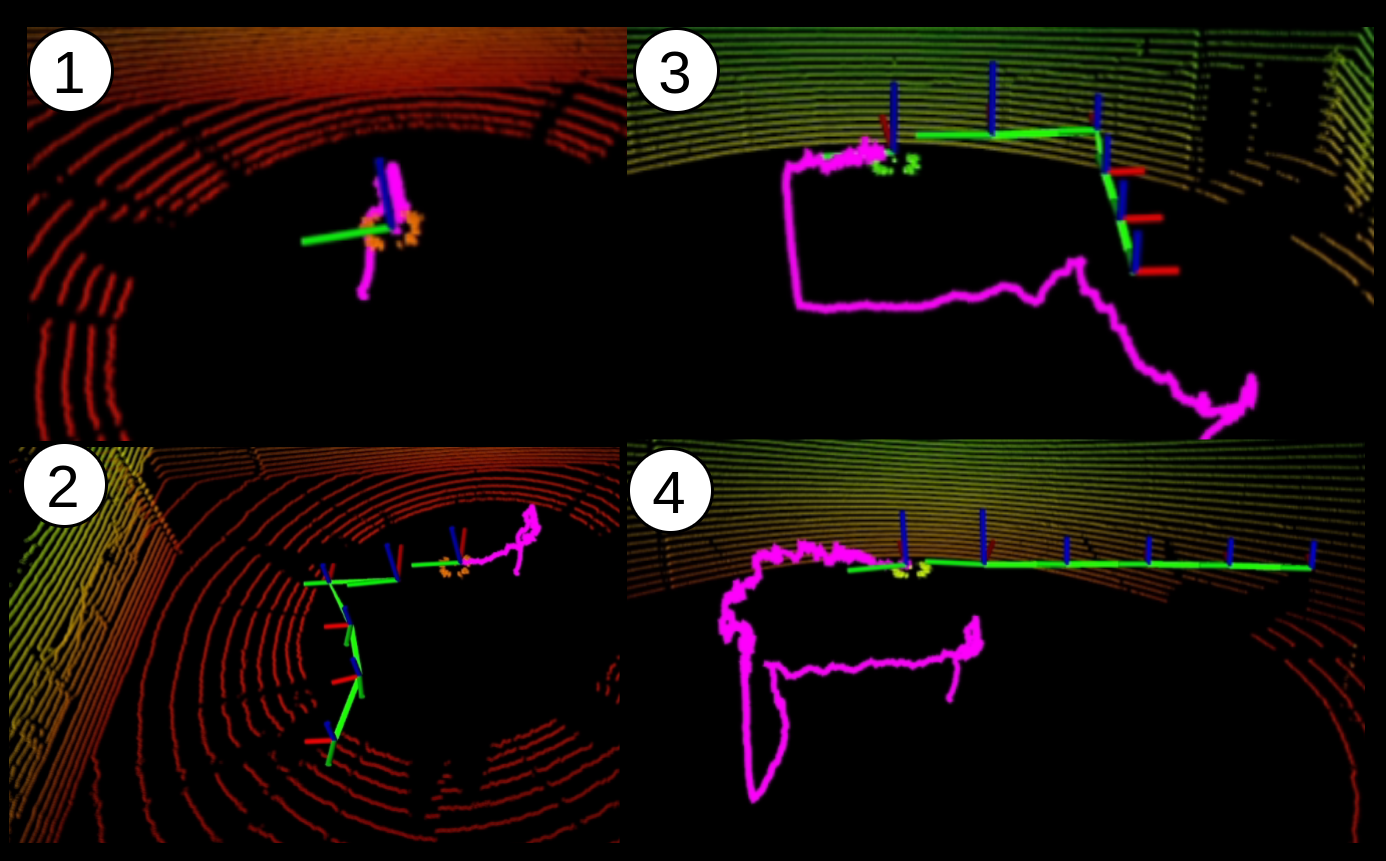}
    \caption{A collage of runtime snapshots taken during the outdoor experimental evaluation of the proposed planning framework. In this run, the aerial platform is tasked to inspect an external corner facade of a building. }
    \label{fig:outdoor_corner}
\end{figure}

Figure.~\ref{fig:outdoor_corner} presents a composite representation of the inspection behaviour when the UAV is tasked to inspect the external corner facade of a building. From Fig.~\ref{fig:outdoor_corner}(2), the predicted inspection route (in \textit{green}) adapts to and tracks the external surface while the UAV is approaching the corner. In Fig.~\ref{fig:outdoor_corner}(3) the planner executes a vertical overlap switch and continues inspection of the external surface with the commanded inspection path, in Fig.~\ref{fig:outdoor_corner}(4), dynamically evolving based on observed surface profile.

\section{Conclusion}\label{sec:conclusions}
This article presents a novel, adaptive to surface characteristics path generator, applicable for MAV operation in large-scale open-pit mines. The proposed scheme consists of two main components, the First-Look inspection plan for surface tracking and the view-pose prediction for surface following with respect to mission and sensor parameters. More specifically, on the sensor parameters, in this work the method considered visual camera 2D footprint for the overlap requirement to maintain data quality measures, which can be adapted to other sensors accordingly. A major merit of the framework is that it doesn't plan over large 3D models of the area, but instead reactively adapts the navigation flight path according to the surface structural morphology. The simulation and experimental results, demonstrate the effectiveness and applicability of the method in such environments.

\bibliography{root}

\end{document}